\documentclass[11pt]{article}

\usepackage[]{acl}
\usepackage{times}
\usepackage{latexsym}
\usepackage[T1]{fontenc}
\usepackage[utf8]{inputenc}
\usepackage{microtype}
\usepackage{CJKutf8}
\usepackage{inconsolata}
\usepackage{graphicx}
\usepackage{CJK}
\usepackage{adjustbox}
\usepackage{pgfplots}
\usepackage{booktabs}
\usepackage{array}
\pgfplotsset{compat=1.18}
\usepackage{amsmath}
\usepackage{xcolor}
\usepackage{tcolorbox}
\tcbuselibrary{skins,breakable}
\usepackage{parskip} 

\newtcolorbox{errorbox}[1]{
  colback=#1!8,
  colframe=#1!40,
  coltitle=black,
  fonttitle=\bfseries,
  sharp corners,
  boxrule=0.5pt,
  left=5pt,
  right=5pt,
  top=5pt,
  bottom=5pt,
  breakable
}

\title{Tag-Aware Structured Text Translation: Towards a Systematic Understanding}

\author{
    Zhanglin Wu\textsuperscript{\rm},
    Hengchao Shang\textsuperscript{\rm},
    Daimeng Wei\textsuperscript{\rm},
    Jiaxin Guo\textsuperscript{\rm},\\
    \bf{Zongyao Li\textsuperscript{\rm},}
    \bf{Tengfei Song\textsuperscript{\rm},}
    \bf{Ning Xie\textsuperscript{\rm},}
    \bf{Weidong Zhang\textsuperscript{\rm}}\\
  \textsuperscript{\rm}Huawei Translation Service Center, Beijing, China\\
   \tt \{wuzhanglin2,shanghengchao,weidaimeng,guojiaxin1,\\
  \tt lizongyao,songtengfei2,zhangweidong17,nicolas.xie\}@huawei.com \\
  }

\begin{document}
\maketitle

\begin{abstract}
Internet texts are replete with format tags that carry structural, semantic, and functional meaning. Current large language model (LLM)-based translation systems struggle to balance translation fluency with tag fidelity when processing tagged text. We argue that resolving this tension requires a systematic approach at three interconnected levels: data synthesis, capability building, and multi-objective alignment. At the data level, we identify and formalize a fundamental trade-off between structural tag diversity and translation naturalness in synthetic data generation; existing methods optimize for one at the expense of the other. We propose a hybrid synthesis strategy (Hy-LST) combining LLM-based synthesis tag method and Two-Stage LLM-based synthesis tag method to produce both diverse and natural tagged data. At the capability level, we decompose tag-aware translation into four sub-tasks of increasing difficulty in a multi-task supervised fine-tuning framework, enabling targeted capability acquisition and knowledge transfer. At the alignment level, we design three complementary reward functions under a group relative policy optimization framework, each targeting a distinct objective (fluency, tag fidelity, and tag-scoped translation quality), and show that joint optimization consistently outperforms single-reward alternatives. Experiments on six language directions (en2zh, en2ja, en2de, en2fr, en2ru, de2fr) demonstrate that each level contributes measurable improvements, and the complete system significantly outperforms existing methods. Qualitative analysis reveals specific error patterns and their mitigation after training with our method.

\end{abstract}

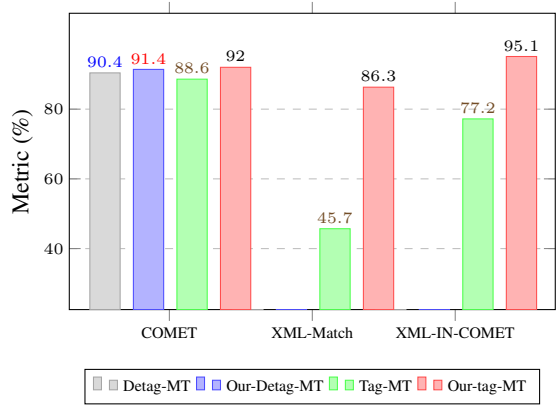
\begin{figure}[ht]
\centering
\hspace{-0.7cm}
\begin{tikzpicture}
    \begin{axis}[
        ybar=0.2cm,
        symbolic x coords={COMET,XML-Match,XML-IN-COMET},
        xtick=data,
        ylabel={\small Metric (\%)},
        legend style={at={(0.5,-0.2)}, anchor=north, legend columns=0.9},
        enlargelimits=0.35,
        width=8cm,
        height=5.5cm,
        bar width=0.4cm,
        ymajorgrids=true,
        grid style=dashed,
        nodes near coords,
        nodes near coords align={vertical},
        ymin=40, ymax=90,
        ytick={40,60,80},
    ]
        \tiny \addplot+[xshift=+0.05cm, fill=gray!30, draw=gray!70] coordinates {(COMET,90.4) (XML-Match,0.0) (XML-IN-COMET,0.0)};
        \tiny \addplot+[xshift=+0.025cm, fill=blue!30, draw=blue!70] coordinates { (COMET,91.4) (XML-Match,0) (XML-IN-COMET,0)} ;
        \tiny \addplot+[xshift=0.0cm, fill=green!30, draw=green!70]coordinates {(COMET,88.6) (XML-Match,45.7)  (XML-IN-COMET,77.2)};
        \tiny \addplot+[xshift=-0.025cm, fill=red!30, draw=red!70] coordinates { (COMET,92.0) (XML-Match,86.3) (XML-IN-COMET,95.1)} ;
        \legend{Detag-MT,Our-Detag-MT,Tag-MT,Our-tag-MT};
    \end{axis}
\end{tikzpicture}
\caption{Our tag-aware system vs. plain SFT: a comparison of detag and tag-aware translation modes in en2zh machine translation.}
\label{fig:mt}
\end{figure}

\section{Introduction}
\label{sec:introduction}

In the context of deep integration between globalization and the internet, cross-lingual information exchange is becoming increasingly frequent. Web pages, documents, and various other forms of online content widely adopt structured tags such as HTML. These tags not only control the presentation format of text but also carry rich semantic connotations and functional value. During machine translation~\cite{vaswani2023attentionneed,sutskever2014sequencesequencelearningneural}, if such tags are not properly handled, it can easily lead to formatting chaos and may further cause semantic distortion and functional failure.

Although large language models (LLMs)~\cite{touvron2023llamaopenefficientfoundation,qwen,deepseekai2024deepseekv3technicalreport} have made significant progress in translating untagged text, they still face considerable challenges when handling tagged text. As shown in Figure~\ref{fig:mt}, after supervised fine-tuning (SFT)~\cite{dong2024abilities} on untagged data (Plain SFT), LLMs exhibit both low tag fidelity and degraded translation quality when translating tagged text. This reveals that vanilla LLMs lack the inherent ability to comprehend tag structures during translation.

Existing approaches fall into two eras. Before the rise of LLMs, two paradigms dominated. The \textbf{detag-and-project method}~\cite{joanis2013transferring,muller2017treatment,zenkel2021automatic} removes tags from the source for translation and then projects them back based on alignment, separating translation from tag handling and causing error propagation. The \textbf{masked tag training method}~\cite{hanneman2020should,elshin2024general} normalizes tags into the translation process, but the sparsity of tag data disrupts attention distributions. Both treat tags as peripheral artifacts rather than integral structures. With LLMs, studies~\cite{dabre2023study} introduce tagged examples as few-shot demonstrations, while others leverage phrase alignment-based synthesis tag method (AST)~\cite{ryu2022data} or LLM-based synthesis tag method (LST)~\cite{dabre2024effective} to synthesize tagged bilingual data for training.

Despite these advances, existing approaches share a critical limitation: they lack a \emph{systematic} understanding of what tag-aware translation requires. We identify three gaps that must be addressed jointly:

\paragraph{Gap 1: Data-level diversity-naturalness trade-off.} The first gap arises at the lowest level of the pipeline: the data itself. Tagged bilingual data is scarce, so synthesis is required. The two dominant methods, AST and LST, both couple tag insertion with target-side alignment, producing natural but structurally rigid training data. Decoupling insertion from translation, as in our Two-Stage LST (LST-2S), increases diversity but degrades translation naturalness. Resolving this trade-off requires a principled hybrid strategy.

\paragraph{Gap 2: Capability-level task decomposition.} Tag-aware translation is not a monolithic task. It requires (a) fluent semantic mapping, (b) tag structure recognition and fidelity, (c) understanding tag semantic scope, and (d) flexible sentence restructuring to accommodate tags. Training all these capabilities under a single objective forces the model to learn competing skills simultaneously, leading to suboptimal performance on each.

\paragraph{Gap 3: Alignment-level multi-objective optimization.} Translation fluency and tag fidelity are inherently competing objectives: preserving tag structure may require rigid sentence alignment, while maximizing fluency may break it. A single reward cannot capture this tension. Multiple reinforcement signals are needed to find the optimal trade-off.

\textbf{Our contributions.} To address these gaps, we propose a systematic framework:
\begin{itemize}
    \item \textbf{Data level:} We formalize the diversity-naturalness trade-off and propose Hy-LST, a hybrid strategy combining LST (high naturalness) with LST-2S (high diversity). To our knowledge, this is the first work to explicitly characterize and address this trade-off.
    \item \textbf{Capability level:} We design a multi-task~\cite{luong2015multi} SFT framework with four tasks of increasing difficulty, enabling knowledge transfer across skills.
    \item \textbf{Alignment level:} We introduce three complementary reward functions under Group Relative Policy Optimization (GRPO)~\cite{shao2024deepseekmath}, targeting fluency, tag fidelity, and tag-scoped translation quality, and demonstrate that joint optimization outperforms both single-reward and no-RL baselines.
\end{itemize}

\section{Method}

\begin{figure}[tbph]
\centering
\includegraphics[height=6cm,width=7.5cm]{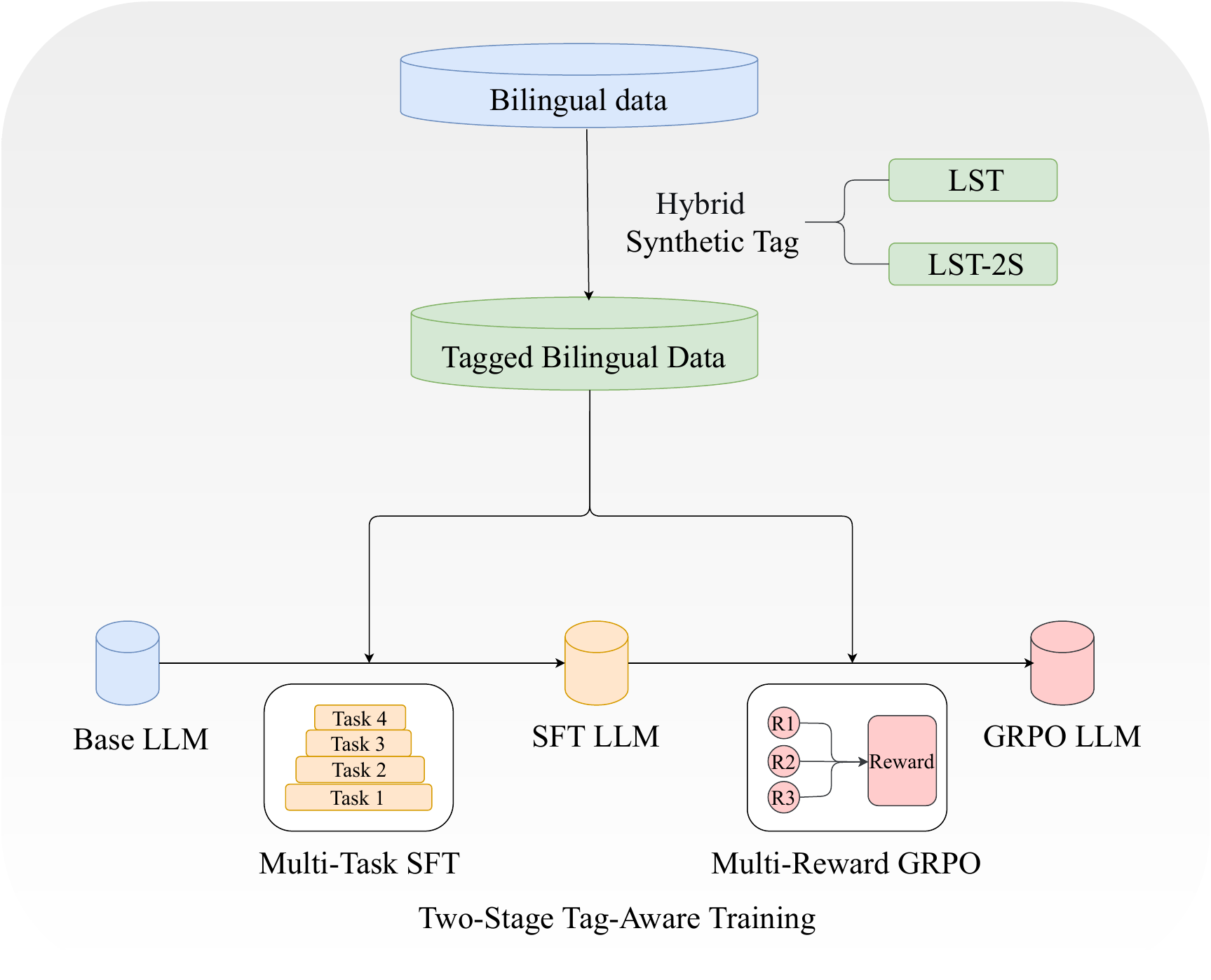}
\caption{Our Proposed systematic multi-level framework.}
\label{fig:method}
\end{figure}

As shown in Figure~\ref{fig:method}, our method addresses the three gaps identified in Section~\ref{sec:introduction} through a systematic multi-level framework.

\begin{figure*}[bpht]
\centering
\includegraphics[height=7cm,width=14cm]{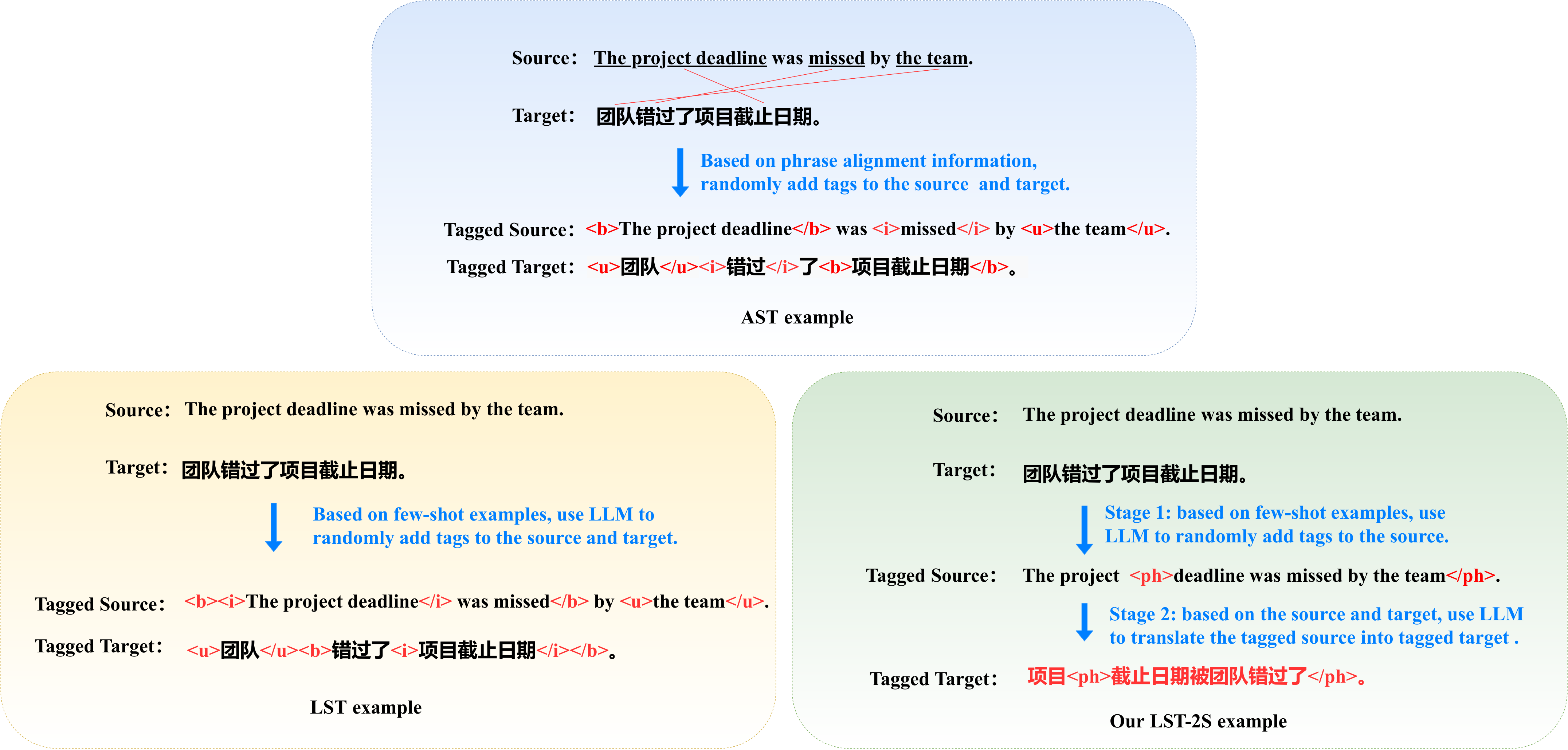}
\caption{Our method (LST-2S) vs. existing methods (AST and LST) for synthesizing en2zh tag examples.}
\label{fig:tag_ex}
\end{figure*}

\subsection{Hybrid Synthetic Tag}
\label{sec:hybrid-syn}

As identified in Gap~1 (Section~\ref{sec:introduction}), tagged bilingual data is scarce, and synthesis is required. The design of a synthesis strategy involves a fundamental trade-off: \textbf{structural diversity} (how varied the tag placements are) versus \textbf{translation naturalness} (how fluent the resulting tagged target text is).

\textbf{Phrase Alignment-Based Synthesis Tag (AST)}~\cite{ryu2022data} obtains bilingual word alignment, expands it to phrase alignment, and wraps tags around aligned phrase pairs. AST achieves high naturalness (tags align with linguistically meaningful units) but low diversity (constrained to alignment boundaries).

\textbf{LLM-based Synthesis Tag (LST)}~\cite{dabre2024effective} uses few-shot prompting to generate tagged bilingual pairs directly. LST produces more natural insertion positions than AST, but still couples source and target tag placement through the generation process, limiting structural variety.

Both methods couple tag insertion with target-side alignment, restricting diversity. To overcome this, we propose a two-stage decoupled strategy:

\textbf{Two-Stage LLM-based Synthesis Tag (LST-2S):} In stage one, an LLM inserts tags randomly into the source text without target constraints, maximizing structural diversity. In stage two, the LLM translates the tagged source into tagged target text, using the original untagged bilingual pair as context. Decoupling allows arbitrary tag placements while enabling the LLM to adapt sentence structures flexibly. Figure~\ref{fig:tag_ex} shows examples.

\textbf{Hybrid LLM-based Synthesis Tag (Hy-LST):} LST-2S achieves higher diversity, but the second-stage translation of arbitrarily tagged sources can degrade fluency. We propose a hybrid strategy combining LST and LST-2S data in equal proportion, balancing LST's naturalness with LST-2S's diversity. The model is thus exposed to both naturally aligned and structurally diverse tag patterns. A detailed experimental comparison of the two methods in terms of tag diversity and translation fluency is presented in Appendix~\ref{subsec:lst-lst2s-comp}.

\subsection{Two-Stage Tag-Aware Training}

The training process is divided into two stages with complementary objectives: SFT builds fundamental tag comprehension, while GRPO fine-tunes the balance between competing objectives.

\subsubsection{Stage 1: Multi-Task SFT}
\label{sec:multi-task-sft}

As identified in Gap~2 (Section~\ref{sec:introduction}), tag-aware translation bundles together multiple competing capabilities. Our solution is to decompose this complex competency into four tasks with increasing difficulty.

Formally, each sentence has two representations: untagged text and tagged text. Let $X$ and $Y$ denote the untagged source and target sentences. Let $T_X$ denote the tagged source (source text with tags inserted), and $T_Y$ the tagged target (target text with tags inserted). The four tasks are ordered by the amount of auxiliary information available.

\begin{align}
&\textbf{Task1:} \quad \mathcal{I}_1 = X, \ \qquad\qquad\; \mathcal{O}_1 = Y \\[4pt]
&\textbf{Task2:} \quad \mathcal{I}_2 = (X, Y, T_X), \quad \mathcal{O}_2 = T_Y \\[4pt]
&\textbf{Task3:} \quad \mathcal{I}_3 = (X, T_X), \qquad \mathcal{O}_3 = T_Y \\[4pt]
&\textbf{Task4:} \quad \mathcal{I}_4 = T_X, \qquad\qquad\; \mathcal{O}_4 = T_Y
\end{align}

The difficulty progression is governed by how much information the model can leverage beyond the tagged source $T_X$. In Task1, the model learns basic translation from $X$ to $Y$, establishing fluency without tag interference. In Task2, the model receives both the untagged bilingual pair $(X, Y)$ and the tagged source $T_X$, providing direct supervision for mapping tagged inputs to tagged outputs. In Task3, the target reference $Y$ is removed, so the model must independently determine sentence restructuring while only having the untagged source $X$ as a guide. In Task4, only the tagged source $T_X$ is given, and the model must simultaneously recognize tag structures, understand their semantic scope, and produce a fluent tagged translation with no external guidance.

By training these tasks jointly, the multi-task framework enables natural knowledge transfer: fluency from Task1 benefits all higher tasks, contextual reasoning from Task2 and Task3 equips the model for the fully independent Task4, and tag awareness acquired across $T_X \rightarrow T_Y$ mappings reinforces understanding throughout the multi-level framework.

\begin{figure}[bpht]
\centering
\includegraphics[height=5cm,width=7cm]{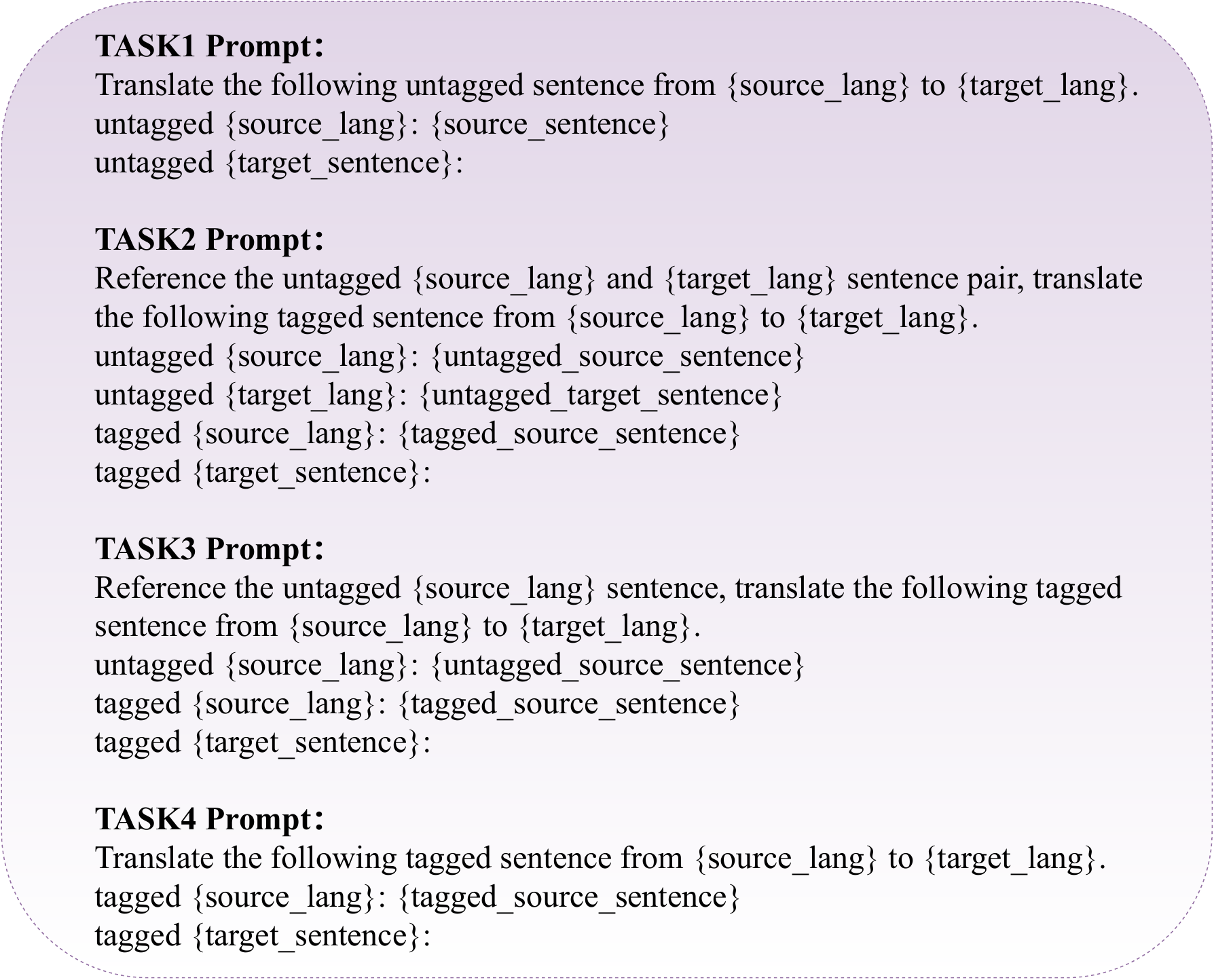}
\caption{Prompt Design for Multi-task SFT.}
\label{fig:prompt}
\end{figure}

\subsubsection{Stage 2: Multi-Reward GRPO}
\label{sec:multi-reward-grpo}

As identified in Gap~3 (Section~\ref{sec:introduction}), translation fluency and tag fidelity are inherently competing objectives, making a single reinforcement learning reward insufficient. To address this, we design three reward functions (Eqs.~\ref{r1}–\ref{r3}) under GRPO and combine them via weighted summation (Eq.~\ref{r-combined}).

\noindent \textbf{Translation Quality Reward ($R_1$):} COMET~\cite{rei2020comet,rei2022comet} is used to evaluate overall fluency on untagged text. To avoid overfitting, COMET-20 is employed during training and COMET-22 during testing.

\begin{equation}
\label{r1}
R_1 = \mathrm{comet}_{20}(X, Y_{mt}, Y_{ref})
\end{equation}

\noindent \textbf{Tag Fidelity Reward ($R_2$):} Exact set matching between the tag set extracted from the model-generated tagged translation $T_Y^{mt}$ and the gold tagged target $T_Y^{ref}$. Let $\mathcal{T}(\cdot)$ denote the set of tags present in a tagged sequence. Then:

\begin{equation}
\label{r2}
R_2 =
\frac{ | \mathcal{T}(T_Y^{mt}) \cap \mathcal{T}(T_Y^{ref}) | }
{ \max( |\mathcal{T}(T_Y^{mt})|, |\mathcal{T}(T_Y^{ref})| ) }
\end{equation}

\noindent \textbf{Tag-Scoped Translation Quality Reward ($R_3$):} Average COMET score across all tag-scoped segments, computed as:

\begin{equation}
\label{r3}
R_3 = \frac{1}{N} \sum_{i=1}^{N} \operatorname{comet}_{20}\bigl( X_i, Y_{\text{mt},i}, Y_{\text{ref},i} \bigr)
\end{equation}

$R_1$ targets fluency, $R_2$ enforces structural fidelity, and $R_3$ ensures tag-scoped translation quality. The combined reward is defined as follows:

\begin{equation}
\label{r-combined}
\scalebox{0.9}{$\displaystyle R_{total} = \frac{\alpha \cdot R_1 + \beta \cdot R_2 + \gamma \cdot R_3}{\alpha + \beta + \gamma}$}
\end{equation}

\noindent where $\alpha$, $\beta$, $\gamma$ control the relative importance of each objective (default $\alpha=\beta=\gamma=1$). This multi-reward formulation jointly optimizes the balance between translation fluency and tag fidelity.

\section{Experiment}

\subsection{Dataset}

\subsubsection{Open-source Dataset}

Our experiments use the multilingual structured document translation dataset~\cite{hashimoto-etal-2019-high}\footnote{\url{https://github.com/salesforce/localization-xml-mt}}, which is multi-way parallel with each English source sentence aligned to translations in Chinese, Japanese, German, French, and Russian; from this dataset we select six translation directions: en2zh, en2ja, en2de, en2fr, en2ru, and de2fr.

For the five English-centric directions, each provides about 100K training instances, 2K development instances, and 2K test instances. Since the test set references are not publicly available, we use the development set for testing instead. To prevent data leakage, we remove any training sample whose source text appears in the test set. For the de2fr direction, we construct bilingual pairs by matching German and French sentences that share the same English source, resulting in approximately 90K training pairs, from which we reserve 2K as the test set. Table~\ref{tab:dataset} summarizes the data statistics.

\begin{table}[!htbp]
\centering
\large
\setlength{\belowcaptionskip}{-0.1cm}
\begin{adjustbox}{width=\columnwidth,center}{
\begin{tabular}{l|cccccc}
\hline
Dataset & \textbf{en2zh}  & \textbf{en2ja} & \textbf{en2de} & \textbf{en2fr} &\textbf{en2ru} &\textbf{de2fr} \\
\hline
Train & 88611 & 90761 & 91333 & 91419 & 88983 & 88514 \\
Test & 2000 & 2000 & 2000 & 2000 & 2000 & 2000 \\
\hline
\end{tabular}
}\end{adjustbox}
\caption{Our extended localization-xml-mt dataset.}
\label{tab:dataset}
\end{table}

\subsubsection{Extended Testset}

The original test set has a low ceiling: only 27\% of samples contain tags, and 82\% of those have just 1-2 tag pairs. To enable meaningful comparison, we augment tags using LST with three different LLMs (Qwen-max~\cite{qwen}, GPT-4o~\cite{hurst2024gpt}, and GLM-4.7). Samples are filtered by XML-MATCH and XML-IN-COMET22-KIWI (thresholds 100 and 80). After expansion, 97\% of samples contain tags and only 44\% have 1-2 pairs. To avoid distributional overlap, we use DeepSeek-v3~\cite{deepseekai2024deepseekv3technicalreport} for training data synthesis, ensuring training and test sets are generated by different LLM families.

\subsubsection{Real-World Testset}
\label{subsec:real-world-subset}

To assess generalization beyond synthetic tags, we create a real-world test set from naturally tagged HTML web pages. The \textbf{English set} includes 2,000 samples crawled from English web pages and is used to evaluate the en2zh direction. By keeping real-world tag structures, each sample offers a complementary evaluation to our extended test set.

\subsection{Evaluation Metrics}
\label{sec:eval-metrics}

We evaluate three aspects: translation quality (using BLEU~\cite{papineni2002bleu} and COMET on untagged text), tag fidelity (via XML-ACC and XML-MATCH~\cite{hashimoto-etal-2019-high}), and tag-scoped translation quality (via our proposed XML-IN-* metrics). Specifically, XML-IN-BLEU extracts text between each tag pair before computing BLEU, avoiding alignment issues in XML-BLEU, while XML-IN-COMET replaces BLEU with COMET for more flexible evaluation (see Appendix~\ref{sec:xml-in-example} for a concrete comparison). For the real-world validation subset without references, we employ COMET22-KIWI~\cite{rei-etal-2022-cometkiwi} to evaluate translation quality.

\subsection{Baselines}
\label{sec:baselines-design}

We choose Qwen2.5-7B as the base LLM. Baselines are organized by research question:

\textbf{How well does a vanilla LLM perform?} \emph{Prompt} (direct translation) and \emph{Few-shot Prompt} (three similar examples retrieved by LABSE~\cite{feng2022language}). \emph{Detag-and-project} removes tags, translates with Plain SFT (i.e., SFT trained on untagged data), and projects back via awesome-align~\cite{dou2021word} and Min-Max Tag Pair Projection~\cite{zenkel2021automatic}.

\textbf{How effective are existing SFT-based methods?} \emph{Masked SFT} (numbered mask tokens), \emph{AST SFT} (phrase alignment synthesis~\cite{och-etal-1999-improved,bird-loper-2004-nltk}), and \emph{LST SFT} (LLM-guided synthesis using DeepSeek-V3 with three few-shot examples per direction).

\textbf{Do our proposed methods provide additional gains?} We construct a progressive ablation: \emph{Hy-LST SFT}, \emph{Multi-Task Hy-LST SFT}, and \emph{+Multi-Reward GRPO}, to quantify the contribution of each component.

\textbf{How do closed-source LLMs compare?} We compare against \emph{Qwen-max}, \emph{GPT-4o}~\cite{hurst2024gpt}, \emph{GLM-4.7}, and \emph{DeepSeek-V3}~\cite{deepseekai2024deepseekv3technicalreport}, evaluated on a real-world testset under zero-shot prompting with tagged input.

\begin{table*}[!htbp]
\centering
\large
\setlength{\belowcaptionskip}{-0.1cm}
\begin{adjustbox}{width=2\columnwidth,center}{
\begin{tabular}{l|cccccc|c}
\hline
\textbf{en2zh} & BLEU  & COMET & XML-ACC & XML-MATCH & XML-IN-BLEU & XML-IN-COMET & Avg.\\
\hline
Prompt & 32.99  & 84.93  & 94.25  & 38.15  & 33.79  & 76.83  & 60.16 \\
Few-shot Prompt & 34.07  & 85.55  & 97.05  & 60.15  & 49.59  & 84.32  & 68.46 \\
Detag-and-project & 55.49  & 90.36  & 95.25  & 73.65  & 56.53  & 90.03  & 76.89 \\
Masked SFT & 55.84  & 90.99  & 99.85  & 79.30  & 77.01  & 94.02  & 82.84 \\
AST SFT & 55.70 & 90.75 & 99.15 & 80.70 & 75.44 & 93.94 & 82.61 \\
LST SFT & 57.16 & 91.33 & 99.95 & 85.10 & 78.22 & 94.48 & 84.37 \\
\hline
Hy-LST SFT & 57.40 & 91.41  & 99.95  & 85.21 & 78.70 & 94.64 & 84.55 \\
Multi-Task Hy-LST SFT & 59.66  & 91.76  & 99.95  & 85.00  & 79.93  & 94.94  & 85.21 \\
 \ \ \ \  +Multi-Reward GRPO & \textbf{60.22}  & \textbf{91.98}  & \textbf{100.00}  & \textbf{86.25}  & \textbf{80.16}  & \textbf{95.10}  & \textbf{85.62}\\
\hline
\end{tabular}
}\end{adjustbox}
\caption{Evaluation results on the en2zh extended testset.}
\label{tab:enzh}
\end{table*}

\subsection{Training Details}

\textbf{SFT:} We use Qwen2.5-7B as the base model and apply LoRA~\cite{gao2024fashiongpt} fine-tuning ($r=8$, $\alpha=16$, all target modules, dropout 0.0~\cite{srivastava2014dropout}) via LlamaFactory~\cite{zheng2024llamafactory}. The warmup ratio is set to 0.1~\cite{fradkin2010effects}, a cosine annealing learning rate scheduler~\cite{liu2022super} is adopted, the learning rate is 1e-4, and gradient accumulation is used to achieve an effective batch size of 32. Training runs up to 3 epochs on 8 GPUs, and the checkpoint with the lowest validation loss is selected.

\noindent \textbf{GRPO:} We use Open-R1~\cite{openr1} with Task4 as the training objective (fastest speed, marginal performance gap compared to other inference modes). Data amount is 1/10 of SFT. 8 GPUs: one runs vLLM~\cite{kwon2023efficient} (7 candidates per source, temperature 1.0), seven handle training (batch size 7 per card, 4-step accumulation). DeepSpeed ZeRO3-offload, bfloat16, learning rate 1e-6, cosine annealing with 5\% warmup, max length 4096. Early stopping after 5 consecutive non-improving evaluations on validation reward; checkpoints are saved every 100 steps.

\section{Results}
\label{sec:results}

We organize our results around the research questions outlined in Section~\ref{sec:baselines-design} with Qwen2.5-7B on the en2zh extended testset (Table~\ref{tab:enzh}).

\subsection{Vanilla LLM Performance}

Without specialized training, vanilla LLMs perform poorly on tag-aware translation. \emph{Prompt} achieves only 60.16 avg. score, with particularly low XML-Match (38.15) and XML-IN-COMET (76.83). \emph{Few-shot Prompt} improves to 68.46 avg., but still lags far behind SFT-based methods. \emph{Detag-and-project} reaches 76.89 avg. by separately handling translation and tag projection, but error propagation limits its XML-Match to 73.65. These results confirm that vanilla LLMs fundamentally lack the ability to comprehend tag structures during translation, and simple prompting or separate-then-project strategies are insufficient.

\subsection{Existing SFT-Based Methods}

All existing SFT-based methods substantially outperform prompting baselines. \emph{Masked SFT} achieves 82.84 avg. (XML-Match 79.30), \emph{AST SFT} reaches 82.61 avg. (XML-Match 80.70), and \emph{LST SFT} leads at 84.37 avg. (XML-Match 85.10). The progression from Masked to AST to LST confirms that synthesis quality matters: LLM-generated tag patterns (LST) produce better tag fidelity than alignment-based (AST) or mask-based approaches. However, even the best existing method, LST SFT, leaves room for improvement across all metrics.

\subsection{Proposed Method Gains}

Our progressive ablation confirms independent contributions from each component:

\textbf{Hy-LST SFT} (84.55 avg.) improves over LST SFT by +0.18 avg. score, primarily through better XML-Match (85.21 vs. 85.10) and XML-IN-COMET (94.64 vs. 94.48), validating that the hybrid synthesis strategy balances structural diversity with translation naturalness.

\textbf{Multi-Task Hy-LST SFT} (85.21 avg.) adds +0.66 over Hy-LST SFT. The largest gains are in BLEU (+2.26) and COMET (+0.35), indicating that knowledge transfer from simpler sub-tasks significantly improves overall translation fluency without sacrificing tag fidelity.

\textbf{+Multi-Reward GRPO} (85.62 avg.) adds another +0.41, concentrated in XML-Match (+1.25), the most challenging tag fidelity metric. Multi-objective alignment through complementary rewards successfully optimizes the fluency-fidelity trade-off that single-reward or no-RL approaches cannot resolve.

\subsection{Comparison with Closed-Source LLMs}

On a real-world test set (Table~\ref{tab:real-world}), our method outperforms all four closed-source LLMs despite using a smaller base model. Qwen-max ranks first among closed-source models (75.36), while DeepSeek-V3 achieves the best translation quality (KIWI 80.45) but poor tag fidelity (55.10), confirming that even advanced LLMs struggle with tag structure without dedicated training.

\begin{table}[!htbp]
\centering
\Huge
\begin{adjustbox}{width=\columnwidth,center}{
\begin{tabular}{l|ccc|c}
\hline
\textbf{en2zh} & KIWI & XML-MATCH  & XML-IN-KIWI & Avg.\\
\hline
Qwen-max & 80.13 & 78.15 & 67.81 & 75.36 \\
GPT-4o & 78.57 & 52.45 & 60.53 & 63.85 \\
GLM-4.7 & 80.07 & 67.35 & 63.95 & 70.46 \\
DeepSeek-V3 & \textbf{80.45} & 55.10 & 60.86 & 65.47 \\
\hline
Prompt & 78.73 & 26.25 & 54.31 & 53.10 \\
LST SFT & 79.04 & 89.30 & 71.04 & 79.79 \\
\hline
Hy-LST SFT & 79.15 & 90.10 & 71.35 & 80.20 \\
Multi-Task Hy-LST SFT & 79.20 & 91.25 & 71.95 & 80.80 \\
 \ \ \ \ +Multi-Reward GRPO & 79.47 & \textbf{92.10} & \textbf{72.43} & \textbf{81.33} \\
\hline
\end{tabular}
}\end{adjustbox}
\caption{Results on the en2zh real-world testset.}
\label{tab:real-world}
\end{table}

\section{Analysis}

\subsection{Hy-LST vs. Other Synthesis Methods}
\label{sec:tag-syn-comp}

\begin{table*}[!htbp]
\centering
\large
\setlength{\belowcaptionskip}{-0.1cm}
\begin{adjustbox}{width=2\columnwidth,center}{
\begin{tabular}{l|cccccc|c}
\hline
\textbf{en2zh} & BLEU  & COMET & XML-ACC & XML-MATCH & XML-IN-BLEU & XML-IN-COMET & Avg.\\
\hline
Plain SFT & 51.54  & 88.57  & 98.45  & 45.70  & 33.66  & 77.23  & 65.86 \\
Clean SFT & 55.29  & 90.60  & 99.90  & 77.20  & 76.18  & 93.34  & 82.09 \\
Masked SFT & 55.84  & 90.99  & 99.85  & 79.30  & 77.01  & 94.02  & 82.84 \\
AST SFT & 55.70 & 90.75 & 99.15 & 80.70 & 75.44 & 93.94 & 82.61 \\
LST SFT & 57.16 & 91.33 & 99.95 & 85.10 & 78.22 & 94.48 & 84.37 \\
LST-2S SFT & 56.06 & 91.13 & 99.90 & 82.90 & 77.66 & 94.14 & 83.63 \\
\hline
LST+LST-2S SFT & \textbf{57.40}  & 91.41  & 99.95  & \textbf{85.21}  & \textbf{78.70}  & 94.64  & \textbf{84.55} \\
AST+LST-2S SFT & 56.45 & 91.15 & 99.90 & 82.60 & 77.73 & 94.20 & 83.67 \\
AST+LST+LST-2S SFT & 57.30  & \textbf{91.42}  & \textbf{100.00}  & 84.40  & 78.56  & \textbf{94.72}  & 84.40 \\
\hline
\end{tabular}
}\end{adjustbox}
\caption{Comparison of tag synthesis methods on en2zh extended testset.}
\label{tab:tag-syn}
\end{table*}

\begin{table*}[!htbp]
\centering
\large
\setlength{\belowcaptionskip}{-0.1cm}
\begin{adjustbox}{width=2\columnwidth,center}{
\begin{tabular}{l|cccccc|c}
\hline
\textbf{en2zh} & BLEU  & COMET & XML-ACC & XML-MATCH & XML-IN-BLEU & XML-IN-COMET & Avg.\\
\hline
Task1 SFT & 51.54  & 88.57  & 98.45  & 45.70  & 33.66  & 77.23  & 65.86 \\
Task1+Task2 SFT & 59.39  & 91.79  & 99.90  & 85.40  & 80.13  & 94.78  & 85.23 \\
Task3 SFT & 58.02  & 91.58  & 100.00  & 85.10  & 79.19  & 94.70  & 84.77 \\
Task4 SFT & 57.40  & 91.41  & 99.95  & 85.21  & 78.70  & 94.64  & 84.55 \\
\hline
Multi-Task SFT (Task1 MT) & 57.63  & 91.40  & -  & -  & -  & -  & - \\
Multi-Task SFT (Task1+Task2 MT) & \textbf{60.18}  & 91.90  & 99.95  & \textbf{85.85}  & \textbf{80.22}  & \textbf{95.04}  & \textbf{85.52} \\
Multi-Task SFT (Task3 MT) & 60.16  & \textbf{91.92}  & \textbf{100.00}  & 85.65  & 79.95  & 94.96  & 85.44 \\
Multi-Task SFT (Task4 MT) & 59.66  & 91.76  & 99.95  & 85.00  & 79.93  & 94.94  & 85.21 \\
\hline
\end{tabular}
}\end{adjustbox}
\caption{Comparison of SFT training tasks on en2zh extended testset.}
\label{tab:multi-task}
\end{table*}

Plain SFT (avg. 65.86) confirms tag exposure is essential. Clean SFT (original tags) reaches 82.09. Among single methods, LST (84.37) outperforms AST (82.61) and LST-2S (83.63). Critically, LST-2S underperforms LST, confirming our motivation: the two-stage decoupling increases diversity but introduces translation quality degradation in second-stage re-translation.

The hybrid LST+LST-2S achieves 84.55, surpassing both individual methods and validating the diversity-naturalness trade-off hypothesis. Adding AST data provides no further gains, suggesting LLM-based synthesis already covers sufficient tag patterns. These results confirm that \textbf{Hy-LST is better than each individual synthetic method}.

\subsection{Multi-Task vs. Single-Task Training}
\label{sec:sft-task-comp}

Task1 (untagged only) performs worst (65.86). Task4 achieves a large jump to 84.55. Task3 (untagged source context) improves to 84.77, and Task2 (bilingual context) reaches 85.23.

Multi-task training brings additional gains across all inference modes. The most notable is Task4 mode: 85.21 vs. 84.55 independently, a +0.66 gain from knowledge transfer. The gap between inference modes narrows from 0.68 to 0.31, indicating that shared representations make the model more robust. These results confirm that \textbf{multi-task training outperforms single-task training}.

\subsection{GRPO Analysis}
\label{sec:grpo-analysis}

\subsubsection{SFT Necessity for GRPO}

To determine whether SFT is essential or GRPO alone can learn tag-aware translation from scratch, we compare three en2zh setups: (1) Multi-Task SFT + Multi-Reward GRPO (our full method), (2) Direct Multi-Reward GRPO without SFT (base model only), and (3) Single-Task SFT (Task4) + Multi-Reward GRPO.

Results in Table~\ref{tab:grpo-skip} show that Direct GRPO (no SFT) scores only 62.18, far below any SFT-based method, confirming that GRPO alone cannot acquire tag-structured translation from scratch—SFT is necessary for basic tag understanding. Moreover, Multi-Task SFT yields better GRPO initialization than Task4-only SFT (85.62 vs. 84.80), showing the multi-task benefits downstream reinforcement learning.

\begin{table*}[!htbp]
\centering
\large
\setlength{\belowcaptionskip}{-0.1cm}
\begin{adjustbox}{width=2\columnwidth,center}{
\begin{tabular}{l|cccccc|c}
\hline
\textbf{en2zh} & BLEU  & COMET & XML-ACC & XML-MATCH & XML-IN-BLEU & XML-IN-COMET & Avg.\\
\hline
Direct Multi-Reward GRPO (no SFT) & 35.42 & 85.67 & 95.10 & 42.30 & 36.15 & 78.42 & 62.18 \\
Task4 SFT + Multi-Reward GRPO & 58.21 & 91.62 & 100.00 & 85.75 & 79.32 & 94.92 & 84.97 \\
Multi-Task SFT + Multi-Reward GRPO & \textbf{60.22} & \textbf{91.98} & \textbf{100.00} & \textbf{86.25} & \textbf{80.16} & \textbf{95.10} & \textbf{85.62} \\
\hline
\end{tabular}
}\end{adjustbox}
\caption{Comparison of SFT initialization for GRPO on en2zh extended testset.}
\label{tab:grpo-skip}
\end{table*}

\subsubsection{Reward Configuration}
\label{sec:grpo-reward-comp}

\begin{table*}[!htbp]
\centering
\large
\setlength{\belowcaptionskip}{-0.1cm}
\begin{adjustbox}{width=2\columnwidth,center}{
\begin{tabular}{l|cccccc|c|c}
\hline
\textbf{en2zh} & BLEU  & COMET & XML-ACC & XML-MATCH & XML-IN-BLEU & XML-IN-COMET & Avg. & Train Cost(h) \\
\hline
Multi-Task Hy-LST SFT & 59.66  & 91.76  & 99.95  & 85.00  & 79.93  & 94.94  & 85.21 & 38 \\
  \ \ \ \  +$R_1$ GRPO & 60.32 & 92.10 & \textbf{100.00} & 84.52 & 79.82 & 94.51 & 85.21 & 44 \\
  \ \ \ \  +$R_2$ GRPO & 59.32 & 91.41 & \textbf{100.00} & 86.34 & 79.63 & 94.76 & 85.24 & 44 \\
  \ \ \ \  +$R_3$ GRPO & 59.46 & 91.45 & 99.95 & 85.36 & 80.29 & 95.24 & 85.29 & 44 \\
  \ \ \ \  +Multi-Reward GRPO (1:1:1) & 60.22  & 91.98  & \textbf{100.00}  & 86.25  & 80.16  & 95.10  & \textbf{85.62} & 44\\
\hline
  \ \ \ \  +Multi-Reward GRPO (2:1:1) & \textbf{60.55} & \textbf{92.15} & \textbf{100.00} & 85.65 & 80.02 & 94.98 & 85.56 & 44\\
  \ \ \ \  +Multi-Reward GRPO (1:2:1) & 60.01 & 91.58 & \textbf{100.00} & \textbf{86.78} & 80.01 & 95.02 & 85.57 & 44\\
  \ \ \ \  +Multi-Reward GRPO (1:1:2) & 60.08 & 91.54 & \textbf{100.00} & 86.24 & \textbf{80.22} & \textbf{95.16} & 85.54 & 44\\
\hline
\end{tabular}
}\end{adjustbox}
\caption{Comparison of GRPO reward configurations on en2zh extended testset.}
\label{tab:grpo}
\end{table*}

\paragraph{Reward specialization.} As Table~\ref{tab:grpo} shows, each single-reward configuration improves only its targeted metric: $R_1$ raises BLEU/COMET but lowers XML-Match; $R_2$ does the opposite; $R_3$ improves XML-IN-* metrics. None achieves a meaningful average gain over the SFT baseline.

\paragraph{Joint optimization.} Combining all three rewards (1:1:1) achieves the highest average score (85.62), surpassing both the SFT baseline (+0.41) and every single-reward configuration. The gain concentrates in XML-Match (+1.25), confirming that multi-reward optimization is essential for the fluency-fidelity trade-off.

\paragraph{Weight sensitivity.} All three skewed configurations (2:1:1, 1:2:1, 1:1:2) outperform single-reward and SFT baselines. Skewing toward $R_1$ improves BLEU (+0.33) at the cost of XML-Match (-0.60); skewing toward $R_2$ shows the reverse. The equal-weight (1:1:1) setting achieves the best average (85.62), though differences among multi-reward variants are marginal (85.54--85.62).

\subsubsection{Cost-Benefit of GRPO}
Multi-Reward GRPO uses only ~6\% of the SFT data but incurs 1.2× training overhead (44 vs.~38 GPU hours). The overall gain is +0.41 avg. (85.21$\to$85.62), concentrated in XML-Match (+1.25), the hardest tag fidelity metric. For tasks where precise tag preservation is critical, this targeted improvement justifies the added cost.

\subsection{Generalization}
\label{sec:generalization}

To assess generalizability, we evaluate across different base models and translation directions.

\paragraph{Cross-model generalization.} On LLaMA-3.1-8B (Table~\ref{tab:llama}), our full method outperforms LST SFT (84.03 vs.\ 82.24 avg.\ on en2zh), with each component's gains replicating those on Qwen2.5-7B, confirming our framework is model-agnostic.

\paragraph{Cross-direction generalization.} Across all five translation directions (Tables~\ref{tab:enja}--\ref{tab:defr}), our method achieves the highest average score with consistent gains over LST SFT, demonstrating effectiveness across diverse language pairs.

\subsection{Error Analysis}
\label{sec:qualitative}

We categorize tag-related errors into four types: omission (dropping a tag), over-translation (inserting a spurious tag), mistranslation (incorrect tag translation), and scope misplacement (incorrect tag span). Table~\ref{tab:error-mitigation} reports error rates across our multi-level framework on the en2zh extended testset. A detailed case study is provided in Appendix~\ref{subsec:error-case-study}.

Without tag training, omission is the dominant error at 48.00\%, followed by scope misplacement at 25.49\%. LST SFT drastically reduces both to 0.25\% and 1.61\%, respectively, making scope misplacement the most frequent remaining error. Our multi-level framework addresses this progressively: Hy-LST SFT lowers scope misplacement to 1.34\%, Multi-Task SFT to 0.89\%, and Multi-Reward GRPO to 0.45\%, while omission drops further to 0.10\%. Overall, our multi-level framework progressively mitigates all four error types, with the most significant gains achieved on omission and scope misplacement.

\begin{table}[!htbp]
\centering
\Huge
\setlength{\belowcaptionskip}{-0.1cm}
\begin{adjustbox}{width=\columnwidth,center}{
\begin{tabular}{lcccc}
\hline
en2zh & Omission$\downarrow$ (\%) & Over-translation$\downarrow$ (\%) & Mistranslation$\downarrow$ (\%) & Scope Misplacement$\downarrow$ (\%) \\
\hline
Prompt & 48.00 & 0.30 & 4.50 & 25.49 \\
Few-shot Prompt &  21.25 &  1.30 & 3.55 & 11.97 \\
Detag-and-project &  4.00 & 0.55 & 0.40 & 4.62 \\
Masked SFT &  3.75 & 0.45 & 0.10 & 4.21 \\
AST SFT &  5.35 & 0.25  & 0.20 & 4.31 \\
LST SFT & 0.25  &  0.10 & 0.05 & 1.61 \\
\hline
Hy-LST SFT & 0.25  &  0.15 & 0.10 & 1.34 \\
Multi-Task Hy-LST SFT & 0.25  &  0.10 & 0.05 & 0.89 \\
\ \ \ \ +Multi-Reward GRPO & \textbf{0.10} & \textbf{0.10} & \textbf{0.05} & \textbf{0.45} \\
\hline
\end{tabular}
}\end{adjustbox}
\caption{Aggregate error rates across our multi-level framework on en2zh extended testset.}
\label{tab:error-mitigation}
\end{table}

\section{Conclusion}

Tag-aware translation faces a fundamental tension between tag fidelity and translation fluency. We address three root causes through a multi-level framework. At the data level, Hy-LST balances structural diversity with translation naturalness. At the capability level, multi-task SFT enables knowledge transfer across sub-tasks of increasing difficulty. At the alignment level, multi-reward GRPO jointly optimizes fluency, tag fidelity, and tag-scoped translation quality. Experiments across six language directions, two base models, and real-world data validate our approach. Each level progressively reduces distinct error patterns: tag omission, tag misplacement, and tag-scoped translation quality degradation. Our method achieves consistent gains over strong baselines.

\section*{Limitations}

Despite these notable advances, several important limitations nonetheless merit careful discussion:

\textbf{Tag type scope.} Our experiments focus exclusively on paired HTML/XML tags. Extending the framework to handle hierarchical formats such as LaTeX, particularly with cross-document references, remains an open direction for future work.

\textbf{Dependence on synthesis LLM quality.} Hy-LST relies on DeepSeek-V3 for data synthesis. The quality and diversity of the generated tagged data are therefore bounded by the capability of this synthesis LLM. If the synthesis LLM produces unnatural tag placements or translation errors during the second stage (LST-2S), these artifacts may propagate to downstream training. Moreover, replacing DeepSeek-V3 with a weaker model could yield substantially lower gains, highlighting the framework's sensitivity to the synthesis LLM's performance.

\textbf{Computational overhead.} Multi-task SFT introduces increased data management complexity by requiring four distinct task formats to be prepared and trained jointly. GRPO adds extra training time—44 vs.~38 hours, a 1.2\texttimes{} overhead. While this overhead remains modest relative to typical LLM training budgets, it may pose practical concerns in resource-constrained settings.

\bibliography{custom}

\appendix

\section{Detailed Comparison of LST and LST-2S}
\label{subsec:lst-lst2s-comp}

We conducted a controlled experiment on the en2zh extended testset using DeepSeek-V3 to compare two methods for tagged data generation: LST and LST-2S. For each sentence in the testset, we generated ten tagged pairs using each method. To evaluate tag diversity, we employed an LLM as a judge approach with Qwen-max, which rated diversity on a scale from 0 to 5. Meanwhile, we measured translation quality on the untagged text using COMET22-KIWI.

The results (Table \ref{tab:lst-lst2s-gen-comp}) show that LST-2S achieves higher tag diversity, with a score of 4.5 compared to 3.7 for LST. However, it yields lower translation quality, scoring 77.17 versus 78.89 for LST. This inverse relationship confirms a clear trade-off between tag diversity and naturalness in translation. These findings are consistent with the discussion in Section~\ref{sec:tag-syn-comp} and motivate the development of the Hy-LST hybrid strategy, which aims to balance both aspects.

\begin{table}[!htbp]
\centering
\large
\setlength{\belowcaptionskip}{-0.1cm}
\begin{adjustbox}{width=\columnwidth,center}{
\begin{tabular}{l|c|c}
\hline
en2zh & Tag Diversity (LLM-as-a-Judge) & Translation Quality (COMET22-KIWI) \\
\hline
LST & 3.7 & 78.89 \\
LST-2S & 4.5 & 77.17 \\
\hline
\end{tabular}
}\end{adjustbox}
\caption{Comparison between LST and LST-2S on the en2zh extended testset.}
\label{tab:lst-lst2s-gen-comp}
\end{table}

\section{XML-IN-BLEU vs. XML-BLEU}
\label{sec:xml-in-example}

This section provides a concrete comparison between XML-IN-BLEU and XML-BLEU metrics described in Section~\ref{sec:eval-metrics}.

\begin{CJK}{UTF8}{gbsn}

\begin{errorbox}{blue}
\underline{\textbf{Source:}} \\
\small <uicontrol>Search</uicontrol> for a <varname>list view </varname> on the <term>fly</term>. \\
\underline{\textbf{Reference:}} \\
\small <term>动态</term><uicontrol>搜索</uicontrol>一个 <varname>列表视图 </varname>。 \\
\underline{\textbf{Translation:}} \\
\small <uicontrol>搜索</uicontrol>一个<varname>列表视图 </varname><term>在动态中</term>。\\
\end{errorbox}

Using the example above, \textbf{XML-BLEU} first splits both the reference and translation into a list of substrings at each tag boundary. For the reference, the resulting list is: \texttt{[“动态”, “搜索”, “一个“, “列表视图”, “。”]}. Similarly, the translation becomes: \texttt{[“搜索”, “一个“, “列表视图”, “在动态中”, “。”]}. XML-BLEU then computes standard BLEU for each corresponding pair of substrings (by position) and averages the scores. Because the order of tag blocks is different (\textit{e.g.}, “动态” appears first in the reference but fourth in the translation list), the per-substring BLEU scores may be very low, leading to a poor overall average. Thus, XML-BLEU is sensitive to the sequence of XML elements; reordering of tags can substantially degrade the metric even if the content within each tag is correct.

In contrast, \textbf{XML-IN-BLEU} ignores the order of tags and instead extracts only the text content inside each pair of matching tags, regardless of where they appear. For the same reference, the extracted contents are: \texttt{[“动态” (from <term>), “搜索” (from <uicontrol>), “列表视图” (from <varname>)]}. For the translation, the extracted contents are: \texttt{[“搜索” (from <uicontrol>), “列表视图” (from <varname>), “在动态中” (from <term>)]}. XML-IN-BLEU then computes standard BLEU for each corresponding \textit{tag type} (matching by tag name, not by position) and averages the scores. Because tag order does not affect the extraction, "动态" and "在动态中" are still compared under <term>, yielding a reasonable BLEU for that tag. As a result, XML-IN-BLEU is robust to word order changes that arise from different tag sequences, focusing instead on the quality of translation within each structural element.

To summarize, XML-BLEU penalizes any mismatch in the order of XML blocks, while XML-IN-BLEU evaluates each tag's content independently. The latter is therefore more suitable for scenarios where structural reordering is allowed or common, such as in flexible document layouts or when translating between languages with different rhetorical structures.

\end{CJK}

\begin{table*}[!htbp]
\centering
\large
\begin{adjustbox}{width=2\columnwidth,center}{
\begin{tabular}{l|cccccc|c}
\hline
\textbf{en2zh} & BLEU  & COMET & XML-ACC & XML-MATCH & XML-IN-BLEU & XML-IN-COMET & Avg.\\
\hline
Prompt & 33.52 & 82.56 & 92.05 & 59.95 & 51.09 & 84.19 & 67.23 \\
Few-shot Prompt & 50.93 & 86.45 & 93.50 & 68.40 & 55.21 & 88.57 & 73.84 \\
Detag-and-project & 55.66 & 90.18 & 94.80 & 72.75 & 57.02 & 89.91 & 76.72 \\
Masked SFT & 56.93 & 90.43 & 99.80 & 78.75 & 77.42 & 93.63 & 82.83 \\
AST SFT & 56.65 & 90.36 & 99.10 & 80.15 & 76.64 & 93.35 & 82.71 \\
LST SFT & 58.88 & 90.93 & 99.85 & 84.70 & 79.46 & 94.36 & 84.70 \\
\hline
Hy-LST SFT & 59.13 & 91.21 & 99.90 & 84.95 & 79.81 & 94.62 & 84.94 \\
Multi-Task Hy-LST SFT & 60.87 & 91.80 & \textbf{100.00} & 86.25 & 81.22 & 94.97 & 85.85 \\
 \ \ \ \ +Multi-Reward GRPO & \textbf{61.33} & \textbf{92.12} & \textbf{100.00} & \textbf{87.40} & \textbf{81.54} & \textbf{95.22} & \textbf{86.27} \\
\hline
\end{tabular}
}\end{adjustbox}
\caption{Evaluation results of LLaMA-3.1-8B on the en2zh extended testset.}
\label{tab:llama}
\end{table*}

\section{Error Case Study}
\label{subsec:error-case-study}

Section~\ref{sec:qualitative} presents Table~\ref{tab:error-mitigation}, which identifies four types of errors: omission, over-translation, mistranslation, and scope misplacement. The table also quantifies how our multi-level framework progressively reduces each error type. To illustrate this progression, we provide four representative examples from the en2zh real-world testset. Each erroneous translation is produced by an intermediate baseline that exhibits a specific failure mode, while the correct translation is generated by our full system, Multi-Task Hy-LST SFT with Multi-Reward GRPO.

These examples reflect the error distribution shown in Table~\ref{tab:error-mitigation}. In the first case, Prompt (with a 48.00\% omission rate) completely drops the \texttt{<b>} tag from the output while retaining its content. This confirms that the model can translate the text but fails to reason about tag structure. In the second case, Masked SFT (0.45\% over-translation) unnecessarily adds an extra \texttt{<b>} tag around \texttt{<varname>}, indicating that mask-based training encourages the model to extend tags beyond their intended boundaries. The third case reveals a more subtle failure: AST SFT (0.20\% mistranslation) renders \texttt{<varname>} as \texttt{<warname>}, a tag name hallucination caused by alignment-based synthesis producing noisy tag label patterns. In the fourth case, LST SFT (1.61\% scope misplacement) preserves the correct set of tags but swaps the content between \texttt{<ph>} and \texttt{<varname>}, which is the most persistent error type. Our full system successfully resolves all four cases.

\begin{CJK}{UTF8}{gbsn}

\begin{errorbox}{red}
\textbf{Omission (48.00\% $\rightarrow$ 0.10\%)} \\
\underline{\textbf{Source:}} \\
\small Add <parmname>\$RecordType</parmname> <b>manu-\\
ally</b> to your <varname>s-control</varname>. \\
\underline{\textbf{Erroneous Translation (Prompt):}} \\
\small 请手动将<parmname>\$RecordType</parmname> 添加到你的<varname>s-control</varname> 中。 \\
\textcolor{green!60!black}{(Prompt completely omits the <b> tag)} \\
\underline{\textbf{Correct Translation (Full System):}} \\
\small <b>手动</b>将<parmname>\$RecordType</parmname> 添加到您的<varname>S-Control</varname>。 \\
\end{errorbox}

\bigskip

\begin{errorbox}{blue}
\textbf{Over-translation (0.45\% $\rightarrow$ 0.10\%)} \\
\underline{\textbf{Source:}} \\
\small <b><title>Scheduling horizon limit</title></b>: Set the number of days to show before the selected <varname>scheduling horizon</varname>. \\
\underline{\textbf{Erroneous (Masked SFT):}} \\
\small <b><title>调度视野限制</title></b>: 设置显示在所选<b><varname>调度视野</varname></b>之前的天数。\\
\textcolor{green!60!black}{(Masked SFT over-translates the <b> tag)} \\
\underline{\textbf{Correct Translation (Full System):}} \\
\small <b><title>计划水平限制</title></b>: 设置在选定的<varname>计划水平</varname>之前显示的天数。 \\
\end{errorbox}

\bigskip

\begin{errorbox}{orange}
\textbf{Mistranslation (0.20\% $\rightarrow$ 0.05\%)} \\
\underline{\textbf{Source:}} \\
\small <varname>welcome emails</varname> are <ph>enabled</ph>\\
\underline{\textbf{Erroneous (AST SFT):}} \\
\small <warname>欢迎邮件</warname>是<ph>启用</ph>的\\
\textcolor{green!60!black}{(AST SFT translates <varname> as <warname>.)}
\underline{\textbf{Correct Translation (Full System):}} \\
\small <varname>欢迎电子邮件</varname>已<ph>启用</ph>\\
\end{errorbox}

\bigskip

\begin{errorbox}{yellow}
\textbf{Scope Misplacement (1.61\% $\rightarrow$ 0.45\%)} \\
\underline{\textbf{Source:}} \\
\small To <ph>add filters</ph> to <varname>dashboards</var name> <p>you created</p>:\\
\underline{\textbf{Erroneous (LST SFT):}} \\
\small 要为<p>你创建的</p><ph>仪表板</ph><varname> 添加过滤器</varname>：\\
\textcolor{green!60!black}{(LST SFT swaps the tag positions of <varname> and <ph>)} \\
\underline{\textbf{Correct Translation (Full System):}} \\
\small 要<ph>添加筛选器</ph>到<p>您创建的</p><var name>仪表板</varname>： \\
\end{errorbox}

\end{CJK}

\section{Per-Direction Evaluation Results}
\label{sec:per-direction-results}

We report extended results for the remaining five translation directions (en2ja, en2de, en2fr, en2ru, de2fr) in Tables~\ref{tab:enja}--\ref{tab:defr}. Across all directions, our full method consistently outperforms all baselines, with multi-task SFT and multi-reward GRPO providing complementary gains.

\begin{table*}[!htbp]
\centering
\large
\setlength{\belowcaptionskip}{-0.1cm}
\begin{adjustbox}{width=2\columnwidth,center}{
\begin{tabular}{l|cccccc|c}
\hline
\textbf{en2ja} & BLEU  & COMET & XML-ACC & XML-MATCH & XML-IN-BLEU & XML-IN-COMET & Avg.\\
\hline
Prompt & 30.97  & 82.04  & 93.30  & 30.90  & 22.62  & 73.86  & 55.62 \\
Few-shot Prompt & 37.20  & 84.82  & 98.00  & 64.15  & 39.14  & 84.28  & 67.93  \\
Detag-and-project & 60.18  & 92.38  & 93.95  & 65.60  & 41.02  & 88.21  & 73.56  \\
Masked SFT & 60.22  & 92.79  & 99.80  & 73.85  & 68.81  & 93.84  & 81.55  \\
AST SFT & 59.97 & 92.27 & 99.15 & 73.45 & 67.82 & 93.67 & 81.06  \\
LST SFT & 62.24 & 93.06 & \textbf{100.00} & 81.45 & 71.58 & 94.79 & 83.85  \\
\hline
Hy-LST SFT & 62.24  & 93.20  & 99.95  & 81.45  & 72.24  & 94.83  & 83.99  \\
Multi-Task Hy-LST SFT & 65.23  & 93.61  & \textbf{100.00}  & 83.50  & 75.01  & 95.20  & 85.43  \\
  \ \ \ \ +Multi-Reward GRPO & \textbf{65.62}  & \textbf{93.76}  & 99.95  & \textbf{83.75}  & \textbf{75.88}  & \textbf{95.47}  & \textbf{85.74} \\
\hline
\end{tabular}
}\end{adjustbox}
\caption{Evaluation results on the en2ja extended testset.}
\label{tab:enja}
\end{table*}

\begin{table*}[!htbp]
\centering
\large
\setlength{\belowcaptionskip}{-0.1cm}
\begin{adjustbox}{width=2\columnwidth,center}{
\begin{tabular}{l|cccccc|c}
\hline
\textbf{en2de} & BLEU  & COMET & XML-ACC & XML-MATCH & XML-IN-BLEU & XML-IN-COMET & Avg.\\
\hline
Prompt & 25.84  & 76.35  & 92.90  & 43.80  & 28.74  & 74.91  & 57.09 \\
Few-shot Prompt & 30.32  & 81.18  & 94.60  & 75.20  & 42.04  & 82.75  & 67.68 \\
Detag-and-project & 51.14  & 89.53  & 97.40  & 81.70  & 50.50  & 89.34  & 76.60 \\
Masked SFT & 48.46  & 88.73  & 99.40  & 87.00  & 63.65  & 90.75  & 79.67 \\
AST SFT & 48.45 & 88.62 & 99.25 & 86.85 & 62.92 & 90.41 & 79.42 \\
LST SFT & 52.73 & 89.53 & \textbf{100.00} & 91.55 & 67.87 & 92.74 & 82.40 \\
\hline
Hy-LST SFT & 52.91  & 89.65  & \textbf{100.00}  & 92.00  & 68.17  & 92.74  & 82.58 \\
Multi-Task Hy-LST SFT & 56.72  & 90.34  & \textbf{100.00} & 91.35  & 71.15  & 93.41  & 83.83 \\
 \ \ \ \ +Multi-Reward GRPO & \textbf{57.35}  & \textbf{90.64}  & \textbf{100.00}  & \textbf{92.25}  & \textbf{71.98}  & \textbf{93.68}  & \textbf{84.32}\\
\hline
\end{tabular}
}\end{adjustbox}
\caption{Evaluation results on the en2de extended testset.}
\label{tab:ende}
\end{table*}

\begin{table*}[!htbp]
\centering
\large
\setlength{\belowcaptionskip}{-0.1cm}
\begin{adjustbox}{width=2\columnwidth,center}{
\begin{tabular}{l|cccccc|c}
\hline
\textbf{en2fr} & BLEU  & COMET & XML-ACC & XML-MATCH & XML-IN-BLEU & XML-IN-COMET & Avg.\\
\hline
Prompt & 36.89  & 80.81  & 92.05  & 46.05  & 33.32  & 74.06  & 60.53 \\
Few-shot Prompt & 43.73  & 81.41  & 96.40  & 71.65  & 47.31  & 82.15  & 70.44 \\
Detag-and-project & 65.05  & 89.98  & 97.90  & 86.20  & 61.37  & 88.84  & 81.56 \\
Masked SFT & 63.84  & 89.29  & 99.55  & 91.50  & 73.24  & 91.44  & 84.81 \\
AST SFT & 63.01 & 88.98 & 98.45 & 90.35 & 72.66 & 90.98 & 84.07 \\
LST SFT & 65.95 & 89.79 & 99.95 & 96.25 & 77.00 & 93.18 & 87.02 \\
\hline
Hy-LST SFT & 66.43  & 90.04  & 99.95  & 95.90  & 77.21  & 93.29  & 87.14 \\
Multi-Task Hy-LST SFT & 68.87  & 90.57  & \textbf{100.00}  & 96.00  & 79.25  & 93.73  & 88.07 \\
  \ \ \ \ +Multi-Reward GRPO & \textbf{69.91}  & \textbf{90.93}  & 99.95  & \textbf{96.52}  & \textbf{79.99}  & \textbf{93.92}  & \textbf{88.54} \\
\hline
\end{tabular}
}\end{adjustbox}
\caption{Evaluation results on the en2fr extended testset.}
\label{tab:enfr}
\end{table*}

\begin{table*}[!htbp]
\centering
\large
\setlength{\belowcaptionskip}{-0.1cm}
\begin{adjustbox}{width=2\columnwidth,center}{
\begin{tabular}{l|cccccc|c}
\hline
\textbf{en2ru} & BLEU  & COMET & XML-ACC & XML-MATCH & XML-IN-BLEU & XML-IN-COMET & Avg.\\
\hline
Prompt & 24.32  & 80.99  & 93.30  & 54.25  & 23.55  & 71.66  & 58.01 \\
Few-shot Prompt & 27.58  & 82.43  & 96.60  & 66.05  & 33.41  & 77.55  & 63.94 \\
Detag-and-project & 44.45  & 90.02  & 96.05  & 80.10  & 44.83  & 86.60  & 73.68 \\
Masked SFT & 42.45  & 89.57  & 99.90  & 86.95  & 62.03  & 89.80  & 78.45 \\
AST SFT & 41.72 & 89.45 & 99.75 & 86.55 & 61.59 & 89.56 & 78.10 \\
LST SFT & 49.48 & 90.66 & 99.95 & 92.5 & 67.24 & 91.89 & 81.95 \\
\hline
Hy-LST SFT & 49.82  & 90.81  & 99.95  & 92.55  & 68.04  & 91.85  & 82.17 \\
Multi-Task Hy-LST SFT & 52.17  & 91.24  & \textbf{100.00}  & 93.65  & 69.81  & 92.42  & 83.22 \\
 \ \ \ \ +Multi-Reward GRPO & \textbf{53.27}  & \textbf{91.60}  & \textbf{100.00}  & \textbf{94.15}  & \textbf{71.03}  & \textbf{92.68}  & \textbf{83.79}\\
\hline
\end{tabular}
}\end{adjustbox}
\caption{Evaluation results on the en2ru extended testset.}
\label{tab:enru}
\end{table*}

\begin{table*}[!htbp]
\centering
\large
\setlength{\belowcaptionskip}{-0.1cm}
\begin{adjustbox}{width=2\columnwidth,center}{
\begin{tabular}{l|cccccc|c}
\hline
\textbf{de2fr} & BLEU  & COMET & XML-ACC & XML-MATCH & XML-IN-BLEU & XML-IN-COMET & Avg.\\
\hline
Prompt & 27.73 & 77.22 & 95.70 & 47.45 & 23.85 & 67.78 & 56.62 \\
Few-shot Prompt & 33.01 & 78.93 & 96.40 & 73.95 & 38.89 & 78.92 & 66.68 \\
Detag-and-project & 54.33 & 87.5 & 97.90 & 88.50 & 52.95 & 85.61 & 77.80 \\
Masked SFT & 53.12 & 86.81 & 99.55 & 93.60 & 64.82 & 88.21 & 81.02 \\
AST SFT & 52.69 & 86.52 & 99.25 & 93.35 & 64.25 & 87.75 & 80.64 \\
LST SFT & 56.23 & 87.31 & 99.95 & 97.35 & 68.60 & 89.62 & 83.18 \\
\hline
Hy-LST SFT & 56.71 & 87.56 & 99.95 & 98.05 & 69.19 & 90.06 & 83.59 \\
Multi-Task Hy-LST SFT & 58.15 & 88.09 & \textbf{100.00} & 98.30 & 70.85 & 90.5 & 84.32 \\
 \ \ \ \ +Multi-Reward GRPO & \textbf{59.19} & \textbf{88.45} & 99.95 & \textbf{98.80} & \textbf{71.57} & \textbf{90.69} & \textbf{84.78} \\
\hline
\end{tabular}
}\end{adjustbox}
\caption{Evaluation results on the de2fr extended testset.}
\label{tab:defr}
\end{table*}

\end{document}